\documentclass[11pt,a4paper]{article}
\usepackage[hyperref]{emnlp2018}
\usepackage{times}
\usepackage{latexsym}
\usepackage{amsmath}
\usepackage{amssymb}
\usepackage{multirow}
\usepackage{url}
\usepackage{adjustbox}

\DeclareMathOperator{\softmax}{\textbf{softmax}}

\aclfinalcopy 

\title{SyntaxSQLNet: Syntax Tree Networks for Complex and Cross-Domain Text-to-SQL Task
}

\author{Tao Yu \quad Michihiro Yasunaga \quad Kai Yang \quad Rui Zhang   \\{\bf Dongxu Wang \quad Zifan Li \quad Dragomir R. Radev}\\
Department of Computer Science, Yale University\\
\scalebox{0.87}[0.9]{{\tt \{tao.yu,\,michihiro.yasunaga,\,k.yang,\,r.zhang,\,dragomir.radev\}@yale.edu}}}

\date{}

\begin{document}
\setlength{\abovedisplayskip}{4pt}
\setlength{\belowdisplayskip}{4pt}

\maketitle
\begin{abstract}

Most existing studies in text-to-SQL tasks do not require generating complex SQL queries with multiple clauses or sub-queries, and generalizing to new, unseen databases.
In this paper we propose SyntaxSQLNet, a syntax tree network to address the \textit{complex} and \textit{cross-domain} text-to-SQL generation task.
SyntaxSQLNet employs a SQL specific syntax tree-based decoder with SQL generation path history and table-aware column attention encoders.
We evaluate SyntaxSQLNet on the \textit{Spider} text-to-SQL task, which contains databases with multiple tables and complex SQL queries with multiple SQL clauses and nested queries.
We use a database split setting where databases in the test set are unseen during training.
Experimental results show that SyntaxSQLNet can handle a significantly greater number of complex SQL examples than prior work, outperforming the previous state-of-the-art model by 7.3\% in exact matching accuracy.
We also show that SyntaxSQLNet can further improve the performance by an additional 7.5\% using a cross-domain augmentation method, resulting in a 14.8\% improvement in total.
To our knowledge, we are the first to study this complex and cross-domain text-to-SQL task.\footnote{Code available at
\href{https://github.com/taoyds/syntaxsql}{https://github.com/taoyds/syntaxsql}}



\end{abstract}

\section{Introduction}

Text-to-SQL task is one of the most important sub-task of semantic parsing in natural language processing (NLP). It maps natural language sentences to corresponding SQL queries.

In recent years, some state-of-the-art methods with Seq2Seq encoder-decoder architectures are able to obtain more than 80\% exact matching accuracy on some complex text-to-SQL benchmarks such as ATIS and GeoQuery. These models seem to have already solved most problems in this area.

\begin{figure}[!t]
    \vspace{-1mm}\hspace{-1mm}
    \centering
    \includegraphics[width=0.46\textwidth]{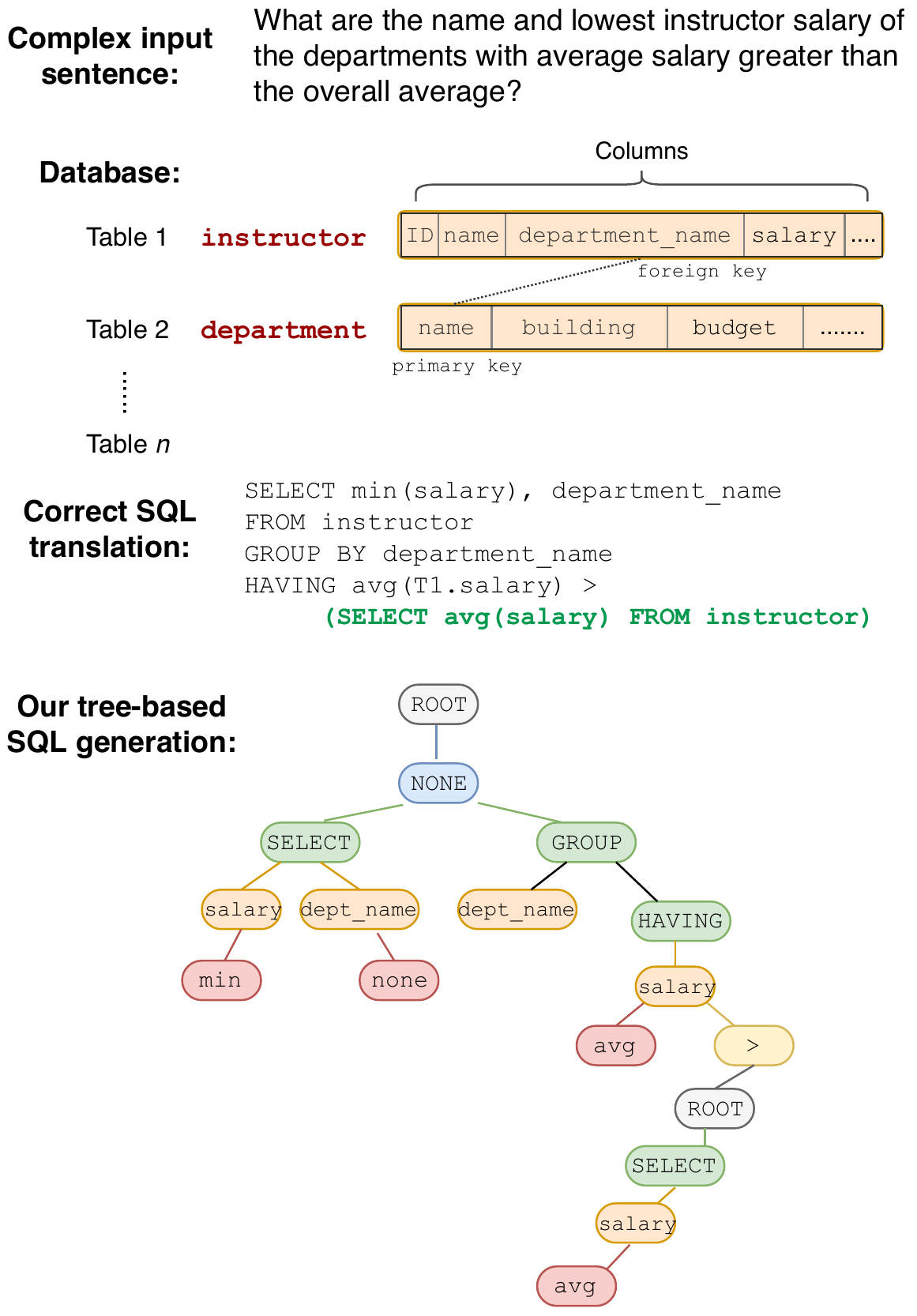}\vspace{-2mm}
    \caption{
    To address the complex text-to-SQL generation task, SyntaxSQLNet employs a tree-based SQL generator. For example, our model can systematically generate a nested query as illustrated above.
    }
\label{fig:task}
\vspace{-3mm}
\end{figure}

However, as \cite{cathy18} show, because of the problematic task definition in the traditional datasets, most of these models just learn to match semantic parsing results, rather than truly learn to understand the meanings of inputs and generalize to new programs and databases.
More specifically, most existing complex text-to-SQL datasets have less than 500 SQL labels. They are expanded by paraphrasing 4-10 questions for each SQL query. Under the standard train and test split \cite{Zettlemoyer05}, most queries in the test set also appear in the train set.
The WikiSQL dataset recently developed by \cite{Zhong2017} is much larger and does use different databases for training and testing, but it only contains very simple SQL queries and database schemas.

To address those issues in the current semantic parsing datasets, 
\citet{Yu&al.18.emnlp.corpus}
have developed a large-scale human labeled text-to-SQL dataset consisting of 10,181  questions, 5,693 unique complex SQL queries, and 200 databases with multiple tables.
They split the dataset into train/dev/test by databases, defining a new \textit{complex} and \textit{cross-domain} text-to-SQL task that requires models to generalize well to both new SQL queries and databases.
The task cannot be solved easily without truly understanding the semantic meanings of the input questions.

In this paper, we propose SyntaxSQLNet, a SQL specific syntax tree network to address the \textit{Spider} task.
Specifically, to generate complex SQL queries with multiple clauses, selections and sub-queries, we develop a syntax tree-based decoder with SQL generation path history.
To make our model learn to generalize to new databases with new tables and columns, we also develop a table-aware column encoder.
Our contributions are as follows: \vspace{-2mm}
\begin{itemize}
\setlength{\itemsep}{-0.5mm}
\setlength{\leftskip}{-4mm}
    \item We propose SQL specific syntax tree networks for the complex and cross-domain text-to-SQL task,
    which is able to solve nested queries on unseen databases.
    We are the first to develop a methodology for this challenging semantic parsing task.
    
    \item We introduce a SQL specific syntax tree-based decoder with SQL path history and table-aware column attention encoders. 
    Even with no hyperparameter tuning, our model can significantly outperform the previous best models, with an 7.3\% boost in exact matching accuracy. Error analysis shows that our model is able to generalize, and solve much more complex (e.g., nested) queries in new databases than prior work.
    \item We also develop a cross-domain data augmentation method to generate more diverse training examples across databases, which further improves the exact matching accuracy by 7.5\%. As a result, our model achieves 27.2\% accuracy, a 14.8\% total improvement compared with the previous best model.
\end{itemize}

\section{Related Work}
\label{sec:rel}

Semantic parsing maps natural language to formal meaning representations. There are a range of representations, such as logic forms and executable programs \cite{zelle96,Zettlemoyer05,wong07,Das10,Liang11,banarescu13,artzi13,Reddy14,Berant14,pasupat2015compositional,herzig18}.

As a sub-task of semantic parsing, the text-to-SQL problem has been studied for decades \cite{warren1982efficient,popescu2003towards,popescu2004modern,li2006constructing,giordani2012translating,wang2017synthesizing}. The methods proposed in the database community \cite{li2014constructing, Yaghmazadeh17} tend to involve hand feature engineering and user interactions with the systems. In this work, we focus on recent neural network-based approaches \cite{Yin15,Zhong2017,Xu2017,Wang2017,iyer17,Izzeddin18,suhr18}. \citet{dong16} introduce a sequence-to-sequence (seq2seq) approach to converting texts to logical forms. Most previous work focuses on a specific table schema.
\citet{Zhong2017} publish the WikiSQL dataset and propose a seq2seq model with reinforcement learning to generate SQL queries.
\citet{Xu2017} further improve the results on the WikiSQL task by using a SQL-sketch based approach employing a sequence-to-set model.
\citet{P18-1068} propose a coarse-to-fine model which achieves the new state-of-the-art performances on several datasets including WikiSQL. Their model first generate a sketch of the target program. Then the model fills in missing details in the sketch.  


Our syntax tree-based decoder is related to recent work that exploits syntax information for code generation tasks \cite{Yin17,RabinovichSK17}.
\citet{Yin17} introduce a neural model that transduces a natural language statement
into an abstract syntax tree (AST). 
While they format the generation process as a seq2seq decoding of rules and tokens, our model uses a sequence-to-set module for each grammar component, and calls them recursively to generate a SQL syntax tree.
Similarly, \citet{RabinovichSK17}
propose abstract syntax networks that use a collection of recursive modules for decoding.
Our model differs from theirs in the following points.
First, we exploit a SQL specific grammar instead of AST.
AST-based models have to predict many non-terminal rules before predicting the terminal tokens, involving more steps. Whereas, our SQL-specific grammar enables direct prediction of SQL tokens.
Second, our model uses different sequence-to-set modules to avoid the ``ordering issue'' \cite{Xu2017} in many code generation tasks. 
Third, different from \cite{RabinovichSK17}, we pass a pre-order traverse of SQL decoding history to each module. 
This provides each module with important dependence information: e.g., if a SQL query has \texttt{GROUP BY}, it is very likely that the grouped column has appeared in \texttt{SELECT} too. 
Finally, instead of sharing parameters across different modules, we train each module separately, because the parameters of different modules could have different converge times.

In addition to the distinction in model design, our work differs from theirs in the data and task definition.
They aim to develop general syntax model for code generation via abstract syntax trees. Instead, we are interested in solving the complex and cross-domain SQL query generation problem; this motivates us to take advantage of SQL specific syntax for decoding, which guides systematic generation of complex SQL queries.


\section{Problem Formulation}
This work aims to tackle the complex text-to-SQL task that involves multiple tables, SQL clauses and nested queries.
Further, we use separate databases for training and testing, aiming to develop models that generalize to new databases.

\paragraph{Dataset.}
We use 
\textit{Spider} \cite{Yu&al.18.emnlp.corpus} \footnote{The Spider task website is at \url{https://yale-lily.github.io/spider}} as the main dataset, which contains 10,181 questions, 5,693 unique complex SQL queries, and 200 databases with multiple tables. 

\paragraph{Task and Challenges.}~\vspace{-1.5mm}
\begin{itemize}
\setlength{\itemsep}{0mm}
\setlength{\leftskip}{-4mm}
    \item The dataset contains a large number of complex SQL labels, which involve more tables, SQL clauses, and nested queries than prior datasets such as WikiSQL. Existing models developed for the WikiSQL task cannot handle those complex SQL queries in the \textit{Spider} dataset.
    
    \item The dataset contains 200 databases ($\sim$138 domains), and different databases are used for training and testing.
    Unlike most previous semantic parsing tasks (e.g., ATIS), this task requires models to generalize to new, unseen databases.
    
\end{itemize}\vspace{-2mm}
In sum, we train and test models on different complex SQL queries from different databases in this task. This aims to ensure that models can make the correct prediction only when they truly understand the meaning of the questions under the given database, rather than by mere memorization.

\section{Methodology}
\label{sec:systems}

Similar to \cite{RabinovichSK17}, our model structures the decoder as a collection of recursive modules.
However, as we discussed in the related work section, we make use of a SQL specific grammar to guide the decoding process, which allows us to take advantage of SQL queries' well-defined structure. Also, modules do not share any parameters so that we train each of them independently.

\begin{figure}[!t]
    \vspace{-1.5mm}\hspace{-1mm}
    \centering
    \includegraphics[width=0.48\textwidth]{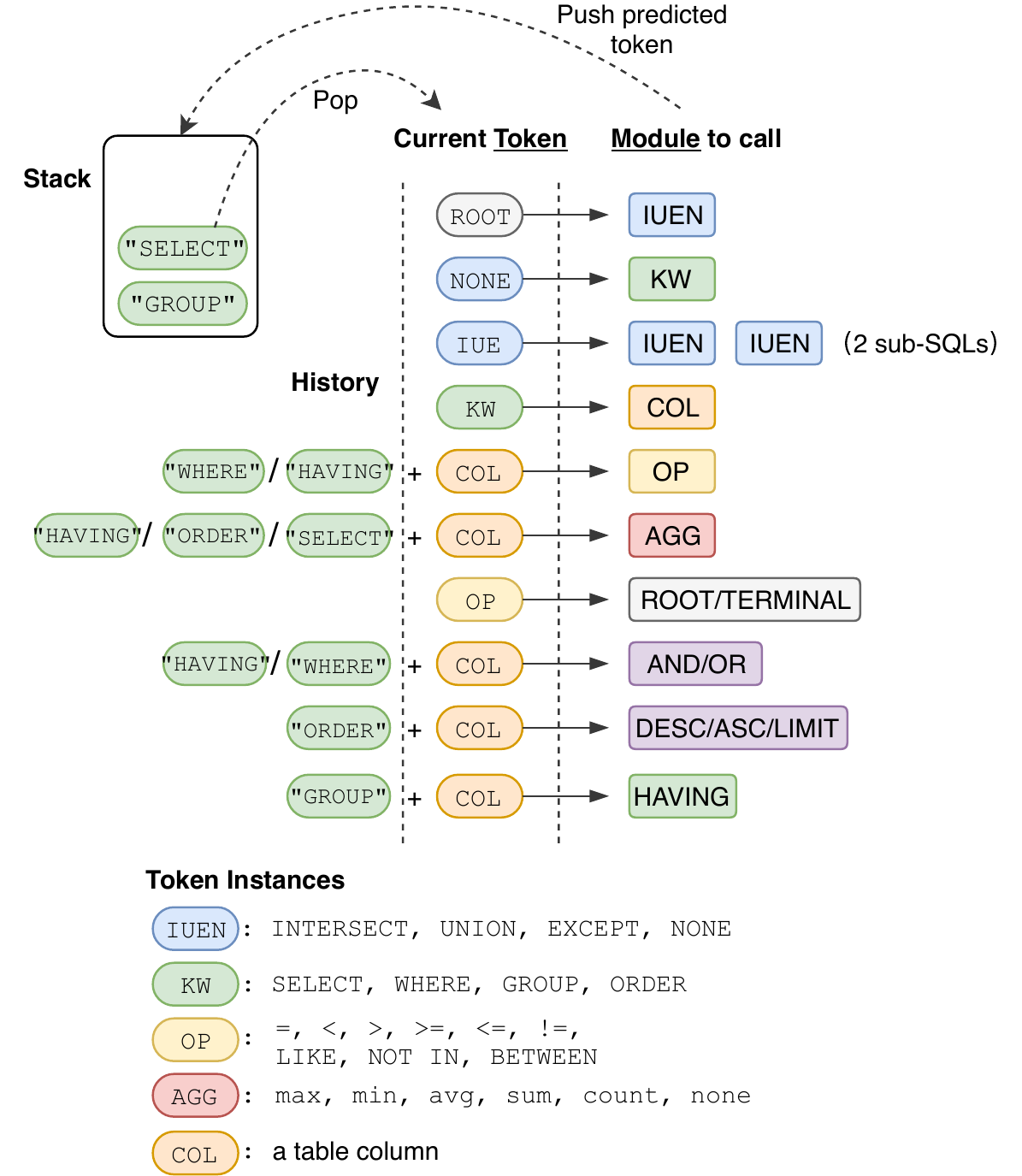}\vspace{-2mm}
    \caption{Our modules and SQL grammar used in decoding process. A round symbol represents a SQL tokens, a table column, etc. A square symbol indicates a module that predicts the next SQL token from its corresponding token instances with the same color.
    }
\label{fig:grammar}
\vspace{-2mm}
\end{figure}

\subsection{Module Overview}
\label{sec:module_overview}
Our model decomposes the SQL decoding process into 9 modules to handle the prediction of different SQL components such as keywords, operators, and columns. We provide the overview in this section and more details in later sections.

Figure \ref{fig:grammar} illustrates our modules and SQL grammar used in decoding process. 
A round symbol represents a SQL token, such as \texttt{SELECT}, \texttt{WHERE}, a table column, etc.
A square symbol indicates a module that predicts the next SQL token from its corresponding token instances with the same color.
Specifically, we have the following modules. \vspace{-2mm}

\begin{itemize}
\setlength{\itemsep}{0mm}
\setlength{\leftskip}{-1mm}
    \item \textbf{IUEN Module}, predicting \texttt{INTERSECT}, \texttt{UNION}, \texttt{EXCEPT}, and \texttt{NONE}, which determines if we need to call itself again to generate nested queries.
    \item \textbf{KW Module}, predicting keywords from \texttt{WHERE}, \texttt{GROUP BY}, and \texttt{ORDER BY}. All queries in our dataset have \texttt{SELECT}.
    \item \textbf{COL Module}, predicting table columns.
    \item \textbf{OP Module}, for $=$, $>$, $<$, $>=$, $<=$, $!\!\!\!=$, \texttt{LIKE}, \texttt{NOT IN}, \texttt{IN}, \texttt{BETWEEN}.
    \item \textbf{AGG Module}, predicting aggregators from \texttt{MAX}, \texttt{MIN}, \texttt{SUM}, \texttt{COUNT}, \texttt{AVG}, and \texttt{NONE}.
    \item \textbf{Root/Terminal Module}, predicting the \texttt{ROOT} of a new subquery or terminal value. It also enables our model to generate nested queries.
    \item \textbf{AND/OR Module}, predicting the presence of AND or OR operator between two conditions.
    \item \textbf{DESC/ASC/LIMIT Module},  predicting the keywords associated with \texttt{ORDER BY}. It is invoked only when \texttt{ORDER BY} is predicted before.
    \item \textbf{HAVING Module}, predicting the presence of \texttt{HAVING} for \texttt{GROUP BY} clause. It is invoked only when \texttt{GROUP BY} is predicted earlier. \vspace{-2mm}
\end{itemize}

\subsection{SQL Grammar}

In order to structure our decoder to generate complex queries, we consider a SQL grammar.
It determines which module to be invoked at each recursive decoding step.
Figure \ref{fig:grammar} illustrates our SQL grammar.
During decoding process, given the current SQL token and the SQL history (the tokens we have gone over to reach the current token), we determine which module to invoke, and predict the next SQL token to generate.

To invoke some modules such as HAVING and OP during decoding, we not only check the type of current token instance but also see whether the type of the previously decoded SQL token is \texttt{GROUP} for HAVING module, and \texttt{WHERE} or \texttt{HAVING} for OP module.

In the grammar, IUEN and Root/Terminal modules are able to generate \texttt{ROOT}, which can activate IUEN module again.
In this way, our model can recursively generate nested subqueries, and can also predict two or more subqueries in queries that have \texttt{EXCEPT}, \texttt{INTERSECT}, and \texttt{UNION}.

\subsection{Input Encoder}

Our inputs of each module consist of three types of information: question, table schema, and current SQL decoding history path.
We encode a question sentence by a bi-directional LSTM, $\rm{BiLSTM}^{\rm{Q}}$.
We encode table schema and history path in the manners described below. 

\subsubsection{Table-Aware Column Representation}
\label{sec:col_emb}
In order to generalize to new databases in testing, it is important to make our model learn to obtain necessary information from a database schema.

SQLNet \cite{Xu2017} encodes this information by running different bi-directional LSTMs over words in each column name, whereas TypeSQL \cite{Yu18} first obtains embedding for each column name by taking the average embedding of the words constituting the column name, and then runs a single biLSTM on the embeddings of all columns in a table. \citet{Yu&al.18.emnlp.corpus} show that the column encoding method of SQLNet outperforms that of TypeSQL in the database split setting, and the result reverses under the example split setting.

While SQLNet and TypeSQL only need the column names as WikiSQL dataset only contains one table per question-SQL pair, \textit{Spider}'s databases contain multiple tables.
To address this setting, we propose to use both table and column names to construct column embeddings. 

Specifically, given a database, for each column, we first get the list for words in its table name, words in its column name, and the type information of the column (string, or number, primary/foreign key), as an initial input of the column. 
Next, like SQLNet, the table-aware column representation of the given column is computed as the final hidden state of a BiLSTM running on top of this sequence. 
This way, the encoding scheme can capture both the global (table names) and local (column names and types) information in the database schema to understand a natural language question in the context of the given database.


We also experimented with a hierarchical table and column encoding, where we first obtain embedding for each table name and then incorporate that information into column encoding. But this encoding method did not perform as well. 

\subsubsection{SQL Decoding History}
\label{sec:hist_path}

In addition to question and column information, we also pass the SQL query's current decoding history as an input to each module. This enables us to use the information of previous decoding states to predict the next SQL token. For example, in Figure \ref{fig:task}, the COL module would be more likely to predict \texttt{salary} in the subquery by considering the path history which contains \texttt{salary} for \texttt{HAVING}, and  \texttt{SELECT} in the main query.

In contract, each module in SQLNet does not consider the previous decoded SQL history. 
Hence, if directly applied to our recursive SQL decoding steps, each module would just predict the same output every time it is invoked.
By passing the SQL history, each module is able to predict a different output according to the history every time it is called during the recursive SQL generation process. Also, the SQL history can improve the performance of each module on long and complex queries because the history helps the model capture the relations between clauses.

Predicted SQL history is used during test decoding.
For training, we first traverse each node in the gold query tree in pre-order to generate gold SQL path history for each training example used in different modules.

\subsubsection{Attention for Input Encoding}
\label{sec:col_emb}


For each module, like SQLNet \cite{Xu2017}, we apply the attention mechanism to encode question representation. We also employs this technique on SQL path history encoding. 
The specific formulas used are described in the next section.










\subsection{Module Details}

Similarly to SQLNet, we employ a sketch-based approach for each module. We apply a sequence-to-set prediction framework introduced by \cite{Xu2017}, to avoid the order issue that happens in seq2seq based models for SQL generation.
For example, in Figure \ref{fig:task}, \texttt{SELECT salary, dept\_name} is the same as \texttt{SELECT dept\_name, salary}.
The traditional seq2seq decoder generates each of them one by one in order; hence the model could get penalized even if the prediction and gold label are the same as sets.
To avoid this problem, SQLNet predicts them together in one step so that their order does not affect the model's training process.
For instance, in Figure \ref{fig:task}, our model invokes the COL module to predict \texttt{salary} and \texttt{dept\_name}, and push to stack at the same time.

However, SQLNet only covers pre-defined SQL sketches, and its modules do not pass information to one another. 
To resolve these problems, SyntaxSQLNet employs a syntax tree-based decoding method that recursively calls different modules based on a SQL grammar. Further, the history of generated SQL tokens is passed through modules, allowing SyntaxSQLNet to keep track of the recursive decoding steps.

We first describe how to compute the conditional embedding $\mathbf{H}_{1/2}$ of an embedding $\mathbf{H}_1$ given another embedding $\mathbf{H}_2$:
$$
\mathbf{H}_{1/2} = \softmax(\mathbf{H}_{1}\mathbf{W} \mathbf{H}_{2}^\top) \mathbf{H}_{1}.
$$
Here $\mathbf{W}$ is a trainable parameter.
Moreover, we get a probability distribution from a given score matrix $\mathbf{U}$ by
$$
\mathcal{P}(\mathbf{U}) = \softmax \left(\mathbf{V} \textbf{tanh}(\mathbf{U}) \right),
$$
where $\mathbf{V}$ is a trainable parameter. 

We denote the hidden states of LSTM on question embeddings, path history, and columns embeddings as $\mathbf{H}_{\textrm{Q}}$, $\mathbf{H}_{\textrm{HS}}$, and $\mathbf{H}_{\textrm{COL}}$ respectively. In addition, we denote the hidden states of LSTM on multiple keywords embeddings and keywords embeddings as $\mathbf{H}_{\textrm{MKW}}$ and $\mathbf{H}_{\textrm{KW}}$ respectively. Finally, we use 
$\mathbf{W}$ to denote trainable parameters that are not shared between modules. The output of each module is computed as follows:

\paragraph{IUEN Module}
In the IUEN module, since only one of the multiple keywords from $\{\texttt{INTERSECT}, \texttt{UNION}, \texttt{EXCEPT}, \texttt{NONE}\}$ will be used, we compute the probabilities by
\begin{gather*}
\resizebox{\hsize}{!}{$
P_{\textrm{IUEN}} = \mathcal{P} \left( \mathbf{W}_{1} \mathbf{H}_{\textrm{Q/MKW}}^\top + \mathbf{W}_{2} \mathbf{H}_{\textrm{HS/MKW}}^\top + \mathbf{W}_{3} \mathbf{H}_{\textrm{MKW}}^\top \right)
$}
\end{gather*}

\paragraph{KW Module}
In the KW module, we first predict the number of keywords in the SQL query and then predict the keywords from $\{\texttt{SELECT}, \texttt{WHERE}, \texttt{GROUP BY}, \texttt{ORDER BY}\}$.
\begin{gather*}
P^{\textrm{num}}_{\textrm{KW}} = \mathcal{P} \left(\mathbf{W}^{\textrm{num}}_{1} {\mathbf{H}^{\textrm{num}}_{\textrm{Q/KW}}}^\top + {\mathbf{W}^{\textrm{num}}_{2} \mathbf{H}^{\textrm{num}}_{\textrm{HS/KW}}}^\top \right)
\\
\resizebox{\hsize}{!}{$
P^{\textrm{val}}_{\textrm{KW}} = \mathcal{P} \left( \mathbf{W}^{\textrm{val}}_{1} {\mathbf{H}^{\textrm{val}}_{\textrm{Q/KW}}}^\top + \mathbf{W}^{\textrm{val}}_{2} {\mathbf{H}^{\textrm{val}}_{\textrm{HS/KW}}}^\top + 
\mathbf{W}^{\textrm{val}}_{3} {\mathbf{H}_{\textrm{KW}}}^\top
\right)$}
\end{gather*}

\paragraph{COL Module}
Similarly, in the COL module, we first predict the number of columns in the SQL query and then predict which ones to use.
\begin{gather*}
P^{\textrm{num}}_{\textrm{COL}} = \mathcal{P} \left(\mathbf{W}^{\textrm{num}}_{1} {\mathbf{H}^{\textrm{num}}_{\textrm{Q/COL}}}^\top + {\mathbf{W}^{\textrm{num}}_{2} \mathbf{H}^{\textrm{num}}_{\textrm{HS/COL}}}^\top \right)
\\
\resizebox{\hsize}{!}{$
P^{\textrm{val}}_{\textrm{COL}} = \mathcal{P} \left( \mathbf{W}^{\textrm{val}}_{1} {\mathbf{H}^{\textrm{val}}_{\textrm{Q/COL}}}^\top + \mathbf{W}^{\textrm{val}}_{2} {\mathbf{H}^{\textrm{val}}_{\textrm{HS/COL}}}^\top + 
\mathbf{W}^{\textrm{val}}_{3} {\mathbf{H}_{\textrm{COL}}}^\top
\right)$}
\end{gather*}

\paragraph{OP Module}
In the OP module, for each predicted column from the COL module that is in the \texttt{WHERE} clause, we first predict the number of operators on it then predict which operators to use from $\{=,\, >,\, <,\, >=,\, <=,\, !\!\!\!\!\!\!\!=,\, \texttt{LIKE},\, \texttt{NOT\!\!\! IN},\, \texttt{IN},\, \texttt{BETWEEN}\}$.
We use $\mathbf{H}_{\textrm{CS}}$ to denote the embedding of one of the predicted columns from the COL module.

\scalebox{0.75}{$
P^{\textrm{num}}_{\textrm{OP}} = \mathcal{P} \left(\mathbf{W}^{\textrm{num}}_{1} {\mathbf{H}^{\textrm{num}}_{\textrm{Q/CS}}}^\top + {\mathbf{W}^{\textrm{num}}_{2} \mathbf{H}^{\textrm{num}}_{\textrm{HS/CS}}}^\top + 
\mathbf{W}^{\textrm{num}}_{3} {\mathbf{H}_{\textrm{CS}}}^\top
\right)
$}

\scalebox{0.75}{$
P^{\textrm{val}}_{\textrm{OP}} = \mathcal{P} \left( \mathbf{W}^{\textrm{val}}_{1} {\mathbf{H}^{\textrm{val}}_{\textrm{Q/CS}}}^\top + \mathbf{W}^{\textrm{val}}_{2} {\mathbf{H}^{\textrm{val}}_{\textrm{HS/CS}}}^\top + 
\mathbf{W}^{\textrm{val}}_{3} {\mathbf{H}_{\textrm{CS}}}^\top
\right)
$}

\paragraph{AGG Module}
In the AGG module, for each predicted column from the COL module, we first predict the number of aggregators on it then predict which aggregators to use from $\{\texttt{MAX},\texttt{MIN},\texttt{SUM},\texttt{COUNT},\texttt{AVG},\texttt{NONE}\}$

\scalebox{0.75}{$
P^{\textrm{num}}_{\textrm{AGG}} = \mathcal{P} \left(\mathbf{W}^{\textrm{num}}_{1} {\mathbf{H}^{\textrm{num}}_{\textrm{Q/CS}}}^\top + {\mathbf{W}^{\textrm{num}}_{2} \mathbf{H}^{\textrm{num}}_{\textrm{HS/CS}}}^\top + 
\mathbf{W}^{\textrm{num}}_{3} {\mathbf{H}_{\textrm{CS}}}^\top
\right)
$}

\scalebox{0.75}{$
P^{\textrm{val}}_{\textrm{AGG}} = \mathcal{P} \left( \mathbf{W}^{\textrm{val}}_{1} {\mathbf{H}^{\textrm{val}}_{\textrm{Q/CS}}}^\top + \mathbf{W}^{\textrm{val}}_{2} {\mathbf{H}^{\textrm{val}}_{\textrm{HS/CS}}}^\top + 
\mathbf{W}^{\textrm{val}}_{3} {\mathbf{H}_{\textrm{CS}}}^\top
\right)
$}

\paragraph{Root/Terminal Module}
To predict nested subqueries, we add a module to predict if there is a new ``ROOT'' after an operator, which allows the model to decode queries recursively. For each predicted column from the COL module that is in the \texttt{WHERE} clause, we first call OP module, and then predict whether the next decoding step is a ``ROOT'' node or a value terminal node by
\begin{gather*}
\resizebox{\hsize}{!}{$
P_{\textrm{RT}} = \mathcal{P} \left( \mathbf{W}_{1} \mathbf{H}_{\textrm{Q/CS}}^\top + \mathbf{W}_{2} \mathbf{H}_{\textrm{HS/CS}}^\top + \mathbf{W}_{3} \mathbf{H}_{\textrm{CS}}^\top \right)
$}
\end{gather*}

\paragraph{AND/OR Module}
For each condition column predicted from the COL module with number bigger than 1, we predict from $\{ \texttt{AND},\texttt{OR}\}$ by
\begin{gather*}
P_{\textrm{AO}} = \mathcal{P} \left( \mathbf{W}_{1} \mathbf{H}_{\textrm{Q}}^\top + \mathbf{W}_{2} \mathbf{H}_{\textrm{HS}}^\top \right)
\end{gather*}

\paragraph{DESC/ASC/LIMIT Module}
In this module, for each predicted column from the COL module that is in the \texttt{ORDER BY} clause, we predict\\ from \scalebox{0.9}[1]{$\{ \texttt{DESC},\texttt{ASC}, \texttt{ DESC LIMIT}, \texttt{ASC LIMIT} \}$} by
\begin{gather*}
\resizebox{\hsize}{!}{$
P_{\textrm{DAL}} = \mathcal{P} \left( \mathbf{W}_{1} \mathbf{H}_{\textrm{Q/CS}}^\top + \mathbf{W}_{2} \mathbf{H}_{\textrm{HS/CS}}^\top + \mathbf{W}_{3} \mathbf{H}_{\textrm{CS}}^\top \right)
$}
\end{gather*}

\paragraph{HAVING Module}
In the HAVING module, for each predicted column from the COL module that is in the \texttt{GROUP BY} clause, we predict whether it is in the \texttt{HAVING} clause by
\begin{gather*}
\resizebox{\hsize}{!}{$
P_{\textrm{HAVING}} = \mathcal{P} \left( \mathbf{W}_{1} \mathbf{H}_{\textrm{Q/CS}}^\top + \mathbf{W}_{2} \mathbf{H}_{\textrm{HS/CS}}^\top + \mathbf{W}_{3} \mathbf{H}_{\textrm{CS}}^\top \right)
$}
\end{gather*}

\subsection{Recursive SQL Generation}
\label{sec:tree_decoder}

The SQL generation process is a process of activating different modules recursively. 
As illustrated in Figure \ref{fig:grammar}, we employ a stack to organize our decoding process.
At each decoding step, we pop one SQL token instance from the stack, and invoke a module based on the grammar to predict the next token instance, and then push the predicted instance into the stack. The decoding process continues until the stack is empty.

More specifically, we initialize a stack with only \texttt{ROOT} at the first decoding step.
At the next step, the stack pops \texttt{ROOT}.
As illustrated in Figure \ref{fig:grammar}, \texttt{ROOT} actives the \texttt{IUEN} module to predict if there is \texttt{EXCEPT}, \texttt{INTERSECT} or \texttt{UNION}. If so, there are two subqueries to be generated in the next step.
If the model predicts \texttt{NONE} instead, it will be pushed into the stack. The stack pops \texttt{NONE} at next step.
For example, in Figure \ref{fig:grammar}, the current popped token is \texttt{SELECT}, which is a instance of keyword (KW) type. It calls the COL module to predict a column name, which will be pushed to the stack.

\subsection{Data Augmentation}
Even though \textit{Spider} already has a significantly larger number of complex queries than existing datasets, the number of training examples for some complex SQL components is still limited.
A widely used way is to conduct data augmentation to generate more training examples automatically.
Many studies \cite{Berant14, iyer17, SuY17a} have shown that data augmentation can bring significant improvement in performance.

In prior work, data augmentation was typically performed within a single domain dataset.
We propose a cross-domain data augmentation method to expand our training data for complex queries.
Cross-domain data augmentation is more difficult than the in-domain setting because question-program pairs tend to have domain specific words and phrases.

To tackle this issue, we first create a list of universal patterns for question-SQL pairs, based on the human labeled pairs from all the different training databases in \textit{Spider}.
To do so, we use a script to remove (and later fill in) all the table \!/\! column names and value tokens in the labeled question-SQL pairs, and then group together the same SQL query patterns.
Consequently, each SQL query pattern has a list of about 5-20 corresponding questions.
In our task, we want to generate more complex training examples.
Thus, we filter out simple SQL query patterns by measuring the length and the number of SQL keywords used.
We obtain about 280 different complex SQL query patterns from over 4,000 SQL labels in the train set of our corpus. We then select the 50 most frequent complex SQL patterns that contain multiple SQL components and nested subqueries. 

After this, we manually edit the selected SQL patterns and their corresponding list of questions to make sure that the table/column/value slots in the questions have one-to-one correspondence to the slots in the corresponding SQL query. For each slot, we also add column type or table information. Thus, for example, columns with string type do not appear in the column slot with integer type during data augmentation (i.e., slot refilling) process.
In this way, our question-SQL patterns are generated based on existing human labeled examples, which ensures that the generated training examples are natural.

Once we have the one-to-one slot mapping between questions and SQL queries, we apply a script that takes a new database schema with type information and generates new question-SQL examples by filling empty slots.
Specifically, for each table in WikiSQL, we first randomly sample 10 question-SQL patterns.
We randomly sample columns from the database schema based on its type: for example, if the slot type in the pattern is ``number'', and then we only sample from columns with ``real'' type in the current table.
We then refill the slots in both the question and SQL query with the selected column names.
Similarly, we also refill table \!/\! value slots.

By this data augmentation method, we finally obtain about 98,000 question and SQL pairs using some WikiSQL databases with one single table.

\section{Experiments}

\subsection{Dataset}
In our experiments, we use \textit{Spider} \cite{Yu&al.18.emnlp.corpus}, a new large-scale human annotated text-to-SQL dataset with complex SQL queries and cross-domain databases. In addition to their originally annotated data, their training split includes 752 queries and 1659 questions from six existing datasets: Restaurants \cite{tang2001using,Popescu03}, GeoQuery \cite{zelle96}, Scholar \cite{iyer17}, Academic \cite{li2014constructing}, Yelp and IMDB \cite{Yaghmazadeh17}.
In total, this dataset consists of 11,840 questions, 6,445 unique complex SQL queries, and 206 databases with multiple tables.
We follow \cite{Yu&al.18.emnlp.corpus}, and use 146, 20, 40 databases for train, development, test, respectively (randomly split).
We also include the question-SQL pair examples generated by our data augmentation method in some experiments.

\subsection{Metrics}
We evaluate our model using SQL Component Matching and Exact Matching proposed by \cite{Yu&al.18.emnlp.corpus}.
To compute the component matching scores, \citet{Yu&al.18.emnlp.corpus} first decompose predicted queries on SQL clauses including \texttt{SELECT}, \texttt{WHERE}, \texttt{GROUP BY}, \texttt{ORDER BY}, and \texttt{KEYWORDS} separately.
After that, they evaluate each predicted clause and the ground truth as bags of several sub-components, and check whether or not these two sets of components match exactly.
Exact matching score is 1 if the model predicts all clauses correctly for a given example.

To better understand model performance on different queries, 
\cite{Yu&al.18.emnlp.corpus} divide SQL queries into 4 levels: easy, medium, hard, extra hard. The definition of difficulty is based on the number of SQL components, selections, and conditions. 

\subsection{Experimental Settings}

Our model is implemented in PyTorch \cite{paszke2017automatic}. We build each module based on the TypeSQL \cite{Yu18} implementation. We use fixed, pre-trained GloVe \cite{pennington14} embeddings for question, SQL history, and schema tokens. For each experiment, the dimension and dropout rate of all hidden layers is set to 120 and 0.3 respectively. We use Adam \cite{Kingma15} with the default hyperparameters for optimization, with a batch size of 64. The same loss functions in \cite{Xu2017} are used for each module. The code is available on \url{https://github.com/taoyds/syntaxsql}. 






\begin{table*}[ht!]
\centering
\scalebox{0.92}{
\begin{tabular}{l|ccccc|c}
\hline
\multirow{2}{*}{Method} & \multicolumn{5}{c|}{Test} & Dev \\ 
& Easy    & Medium  & Hard   & Extra Hard & All & All   \\ \hline
Seq2Seq                         & 11.9\% & 1.9\% & 1.3\% & 0.5\% & 3.7\% & 1.9\%  \\
Seq2Seq+Attention  & 14.9\% & 2.5\% & 2.0\% & 1.1\% & 4.8\% & 1.8\%   \\
Seq2Seq+Copying                 & 15.4\% & 3.4\% & 2.0\% & 1.1\% & 5.3\% & 4.1\% \\
SQLNet          & 26.2\% & 12.6\% & 6.6\% & 1.3\% & 12.4\% & 10.9\%   \\
TypeSQL             & 19.6\% & 7.6\% & 3.8\% & 0.8\% & 8.2\% & 8.0\%  \\\hline
SyntaxSQLNet	          & \textbf{48.0\%} & \textbf{27.0\%}  & \textbf{24.3\%} & \textbf{4.6\%} & \textbf{27.2\%}  & \textbf{24.8\%} \\
-augment	                    & 38.6\%  & 17.6\% & 16.3\% & 4.9\%  & 19.7\% & 18.9\% \\
-table -augment & 37.5\% & 13.5\% & 12.4\% & 1.3\% &  16.4\% & 15.9\%\\
-history -table -augment ~~~  & 18.1\% & 7.0\% & 0.2\% & 0.0\%  & 6.8\% & 6.1\% \\
\hline
\end{tabular}}
\vspace{-2mm}
\caption{Accuracy of Exact Matching on SQL queries with different hardness levels.}
\label{tab:results}
\vspace{-1mm}
\end{table*}

\begin{table*}[ht!]
\centering
\scalebox{0.92}{
\begin{tabular}{lcccccc}
\hline
 Method           & \texttt{SELECT} & \texttt{WHERE} & \texttt{GROUP BY} & \texttt{ORDER BY} & \texttt{KEYWORDS} \\\hline
Seq2Seq           & 13.0\% & 1.5\% & 3.3\% & 5.3\% & 8.7\%  \\
Seq2Seq+Attention & 13.6\% & 3.1\% & 3.6\% & 9.9\% & 9.9\%  \\
Seq2Seq+Copying   & 12.0\% & 3.1\% & 5.3\% & 5.8\% & 7.3\%  \\
SQLNet            & 44.5\% & 19.8\% & 29.5\% & 48.8\% & 64.0\% \\
TypeSQL           & 36.4\% & 16.0\% & 17.2\% & 47.7\% & 66.2\% \\ \hline
SyntaxSQLNet      & \textbf{62.5\%} & \textbf{34.8\%} & \textbf{55.6\%} & \textbf{60.9\%} & 69.6\% \\
-augment	        & 53.9\% & 24.5\% & 44.4\% & 49.5\% & \textbf{71.3\%} \\
-table -augment 	    & 48.9\% & 20.1\% & 36.3\% & 46.8\% & 69.7\% \\
-history -table -augment ~~~	    & 26.7\%	& 14.6\%	  & 11.8\%	& 34.9\%	    & 64.6\% \\\hline
\end{tabular}}
\vspace{-2mm}
\caption{F1 scores of Component Matching on all SQL queries on Test set.}
\label{tab:results_component}
\vspace{-3mm}
\end{table*}

\section{Results and Discussion}

Table \ref{tab:results} presents SyntaxSQLNet's dev and test results compared to previous state-of-the-art models on the \textit{Spider} dataset with database splitting. Our model with SQL history and data augmentation achieves 27.2\% exact matching on all SQL queries, which is about 15\% absolute increase compared to the previous best models, SQLNet and TypeSQL. 

\subsection{Comparison to Existing Methods}

Even though our individual modules are similar to SQLNet and TypeSQL, 
our syntax-aware decoder
allows the modules to generate complex SQL queries in a recursive manner based on the SQL grammar. 
In addition, by incorporating the SQL decoding history into modules during the decoding process, SyntaxSQL achieves a significant gain in exact matching for queries of all hardness levels.
Specifically, even without our data augmentation technique, SyntaxSQLNet outperforms the previous best, SQLNet, by 7.3\%.
This result suggests that the syntax and history information is beneficial for this complex text-to-SQL task.

Moreover, the tree-based decoder enables SyntaxSQLNet to systematically generate nested queries, boosting the performance for Hard/Extra Hard.
As Table \ref{tab:results} shows, SyntaxSQLNet achieves particularly high scores 24.3\% and 4.6\% for Hard and Extra Hard, which contain nested queries. 
The Seq2Seq models suffer from generating ungrammatical queries, yielding very low exact matching accuracy on Hard and Extra Hard SQL queries. In contrast, our model generates valid SQL queries by enforcing the syntax.

For the detailed component matching results in Table \ref{tab:results_component}, our model consistently outperforms other previous work by significant margins. Specifically, our model improve F1 score for most of the SQL components by more than 10\%.

\subsection{Ablation Study}
In order to understand the techniques that are responsible for the performance of our model, we perform an ablation study where we remove one of the proposed techniques from our model at a time. The exact match scores are shown in the same tables as other previous models.

\paragraph{Data Augmentation}
Our model's exact matching performance on all queries drops 7.5\% by excluding data augmentation technique. 
This drop is particularly large for \texttt{GROUP BY} and \texttt{ORDER BY} components (Table \ref{tab:results_component}), for which the original \textit{Spider} dataset has a relatively small number of training examples.
Our cross-domain data augmentation technique provides significantly more examples for column prediction (especially under \texttt{GROUP BY} and \texttt{ORDER BY} clauses), which greatly benefits the overall model performance.


\paragraph{Column Encoding}
To see how our table-aware column encoding affects performance of our model, we also report the model's result without using table information for our column encoding. After excluding the table embedding from column embeddings, the test performance further goes down by 3.3\%. 
This drop is especially large for Medium/Hard SQL queries, where the correct column prediction is a key.
Additionally, in Table \ref{tab:results_component}, the model's performance on \texttt{GROUP BY} component decreases dramatically because it is hard to predict group-by columns correctly without table information (e.g. multiple different tables may have a column of the same name "id" in the database).
This result shows that the table-aware encoding is important to predict the correct columns in unseen, complex databases (with many foreign keys).

\paragraph{SQL Decoding History}
In order to gain more insight into how our SQL decoding history addresses complex SQL, we report our model's performance without SQL path history.
As shown in the Table \ref{tab:results}, the model's performance drops about 9.6\% on exacting matching metric without considering the previous decoding states in each decoding state. More importantly, its performance on hard and extra hard SQL queries decreases to 0\%. This indicates that our model is able to predict nested queries thanks to the SQL decoding history.

\subsection{Error Analysis and Future Work}
The most common errors are from column prediction. Future work may include developing a database schema encoder that can capture relationships among columns and foreign keys in the database more effectively. Other common errors include incorrect prediction of SQL skeleton structures, aggregators and operators. 

There are also a few limitations in our model. For example, SyntaxSQLNet first predicts all the column names in the SQL query, and then chooses tables to generate the \texttt{FROM} clause based on the selected columns. Suppose the natural language input is ``return the stadium name and the number of concerts held in each stadium." The SQL query predicted by SyntaxSQLNet is
\begin{itemize}
    \item[] \texttt{SELECT count(*), name FROM stadium GROUP BY stadium\_id}
\end{itemize}
While the correct answer is
\begin{itemize}
    \item[] \texttt{SELECT T2.name,  count(*) FROM concert AS T1 JOIN stadium AS T2 ON T1.stadium\_id  =  T2.stadium\_id GROUP BY T1.stadium\_id}
\end{itemize}
Even though SyntaxSQLNet predicts all column names and keywords correctly, its deterministic \texttt{FROM} clause generation method fails to join tables ("concert" and "stadium" in this case) together. One possible solution is to predict table names in the \texttt{FROM} clause by considering the relations among tables in the database.

\section{Conclusion}
\label{sec:conclusion}

In this paper, we presented a syntax tree-based model to address complex and cross-domain text-to-SQL task. Utilizing a SQL specific syntax decoder, as well as SQL path history and table-aware column attention encoders, our model outperforms previous work by a significant margin. The ablation study demonstrates that our proposed techniques are able to predict nested, complex SQL queries correctly even for unseen databases.

\section*{Acknowledgement}
We thank Graham Neubig, Tianze Shi, and three anonymous reviewers for their helpful feedback and discussion on this work.

\bibliographystyle{acl_natbib_nourl}
\bibliography{emnlp2018}
\end{document}